\documentclass[10pt,twocolumn,letterpaper]{article}

\usepackage[pagenumbers]{wacv}
\usepackage{amsmath,amssymb,amsfonts}
\usepackage{booktabs}
\usepackage{array}
\usepackage{tabularx}
\usepackage{enumitem}
\usepackage{algorithm}
\usepackage{algpseudocode}

\newcommand{\method}{RAHF-Net}
\newcommand{\mrgb}{\mathbf{F}_{\mathrm{rgb}}}
\newcommand{\mlidar}{\mathbf{F}_{\mathrm{lidar}}}
\newcommand{\mfused}{\mathbf{F}_{\mathrm{fused}}}
\newcommand{\mnear}{\mathbf{F}_{\mathrm{near}}}
\newcommand{\mfar}{\mathbf{F}_{\mathrm{far}}}
\newcommand{\vcyl}{\mathbf{V}_{\mathrm{cyl}}}
\newcommand{\maskm}{\mathbf{M}}
\newcommand{\sig}{\sigma}

\definecolor{wacvblue}{rgb}{0.21,0.49,0.74}
\usepackage[breaklinks,colorlinks,allcolors=wacvblue]{hyperref}

\title{RAHF-Net: Range-Aware Heterogeneous Fusion for\\
LiDAR-Assisted RGB Semantic Segmentation}

\author{%
R.D. Lin$^{1}$ \quad Pengcheng Weng$^{1}$ \quad Xilei Han$^{1}$ \quad Ge Wang$^{2}$ \quad Han Ding$^{2}$ \quad Fei Wang$^{1,\dagger}$\\[0.6em]
{\small $^1$School of Software Engineering, Xi'an Jiaotong University, China}\\
{\small $^2$School of Computer Science and Technology, Xi'an Jiaotong University, China}\\[0.2em]
{\small \texttt{rdlin@stu.xjtu.edu.cn} \quad \texttt{\{dinghan,feynmanw\}@xjtu.edu.cn}}\\[0.2em]
{\small $^\dagger$Corresponding author: \texttt{feynmanw@xjtu.edu.cn}}\\[0.2em]
{\small Code: \href{https://github.com/rdlin118/RAHFNet}{\texttt{github.com/rdlin118/RAHFNet}}}
}

\begin{document}
\maketitle
\begin{abstract}

This paper studies LiDAR-assisted RGB semantic segmentation, where LiDAR point-cloud features improve dense pixel-wise predictions in the RGB image plane rather than 3D point labels or detections. Existing 2D fusion methods often project point clouds into shallow depth/range maps and apply spatially uniform fusion, discarding rich 3D structure and ignoring the range-dependent reliability of outdoor LiDAR. We propose RAHF-Net, a Range-Aware Heterogeneous Fusion Network. Its Cylindrical-to-Perspective (C2P) module projects pretrained cylindrical voxel features into the image plane, preserving structured 3D semantic priors. A Range-Aware Heterogeneous Fusion (RAHF) module then uses a soft depth-derived range mask to perform LiDAR-guided geometric enhancement in reliable near-field regions and visual-context enhancement in sparse far-field regions. On KITTI-360, RAHF-Net achieves 65.76 mIoU, outperforming representative RGB-only and RGB-LiDAR baselines, including CMX by 1.45 points. Experiments on DELIVER further demonstrate improved robustness under adverse weather, illumination degradation, and sensor corruption.

\end{abstract}

\section{Introduction}
\label{sec:intro}

\begin{figure}[t]
    \centering
    \includegraphics[width=1\linewidth]{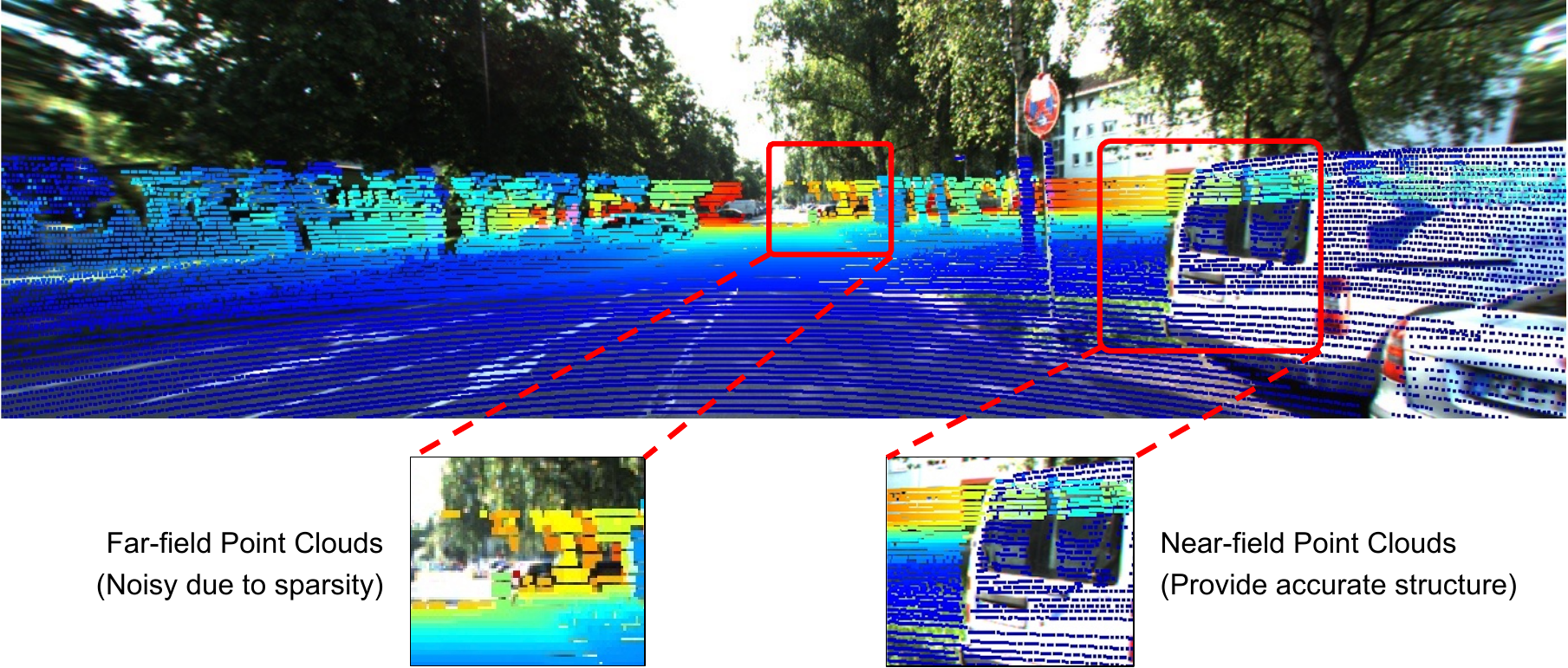}
    \caption{Range-heterogeneity of LiDAR point clouds in driving scenes. Near-field points are dense and reliable, while far-field points are sparse and more likely to introduce noise in multi-modal fusion.}
    \label{fig:range}
\end{figure}

{
Semantic segmentation is a core perception task for autonomous driving, where dense pixel-level scene understanding supports safe planning, obstacle avoidance, and decision making~\cite{guo2018review,zhang2026study}. Modern RGB-based segmentation models have achieved strong performance through fully convolutional networks, encoder-decoder designs, and transformer architectures~\cite{long2015fully,ronneberger2015u,cheng2022masked,xie2021segformer,liu2021swin}. Nevertheless, camera-only perception remains physically tied to the imaging condition. In night scenes, rain, fog, over-exposure, motion blur, or shadowed regions, visual texture and color can become unreliable, causing severe degradation for safety-critical categories such as pedestrians, traffic signs, and vehicles~\cite{9729634,zhou2019automated,zhang2026study}. LiDAR provides complementary depth and geometric structure that is less sensitive to illumination, making RGB-LiDAR fusion an important direction for robust driving-scene segmentation~\cite{8917447,zhang2023cmx,li2023mseg3d,kong2025multi}.

In this paper, we focus on a specific and relatively underexplored setting: using LiDAR point-cloud features to improve RGB image-plane semantic segmentation. The output of the model is a dense 2D semantic label map aligned with the camera image, rather than point-wise 3D segmentation, 3D object detection, or semantic scene completion. This task setting is different from many recent LiDAR-camera perception studies that use RGB to enhance 3D point-cloud prediction. Here, LiDAR acts as an auxiliary geometric source for the RGB segmentation branch, so the central challenge is how to transfer 3D point-cloud knowledge into the 2D image plane without introducing unreliable projected noise.

Despite this complementarity, effective RGB-LiDAR fusion for 2D semantic segmentation remains challenging. A common pipeline first projects LiDAR observations into the image plane as depth maps, range maps, or sparse geometric channels, and then fuses them with image features using concatenation, convolution, attention, or token exchange~\cite{hazirbas2016fusenet,vora2020pointpainting,wang2022multimodal,zhang2023cmx,broedermann2023hrfuser}. Such a pipeline is convenient because it converts multi-sensor alignment into a 2D feature-fusion problem. However, it also introduces two limitations that are especially important in autonomous driving. First, projection compresses a structured 3D signal into a shallow 2D map. The voxel topology, radial density variation, and high-level semantic geometry learned by modern LiDAR encoders are largely discarded before fusion. Second, most fusion modules share parameters spatially over the whole image. This assumes that LiDAR is equally informative everywhere, which is inconsistent with outdoor sensing physics.

The reliability of LiDAR is strongly range-dependent. As illustrated in Figure~\ref{fig:range}, near-field points are dense and can provide accurate object boundaries, making them valuable for correcting RGB ambiguity under shadows, reflections, or strong illumination. In contrast, far-field points are sparse, incomplete, and more vulnerable to calibration error or quantization noise. Forcing the same RGB-LiDAR interaction in both regions may therefore underuse reliable near-field geometry while injecting noisy far-field features into visually coherent regions. This observation suggests that RGB-LiDAR segmentation should be both representation-aware and range-aware: it should transfer richer 3D semantic features into the image branch, and it should adapt the fusion behavior to the physical reliability of LiDAR across distance.

Motivated by this analysis, we propose \method, a Range-Aware Heterogeneous Fusion Network for LiDAR-assisted RGB image-plane semantic segmentation. Instead of treating LiDAR as only a projected depth channel, \method\ reuses a pretrained cylindrical 3D encoder~\cite{zhu2021cylindrical,zhou2020cylinder3d} to extract high-level voxel features that preserve the radial structure and density pattern of driving-scene point clouds. We then design a Cylindrical-to-Perspective (C2P) projection module that maps non-empty cylindrical voxel features into the perspective image plane, enabling pixel-level 2D segmentation to benefit from structured 3D priors. On top of this aligned representation, we introduce a Range-Aware Heterogeneous Fusion (RAHF) module. RAHF generates a soft range mask from projected depth and decomposes feature interaction into two complementary branches: a near-field branch that uses LiDAR geometry to refine boundaries, and a far-field branch that suppresses sparse LiDAR cues while strengthening global visual context.

This design differs from uniform multimodal fusion in both what is fused and how it is fused. C2P transfers high-level cylindrical features rather than shallow depth values, reducing the information loss caused by 3D-to-2D compression. RAHF then performs distance-adaptive fusion rather than global feature mixing, allowing the model to call on the most reliable modality according to spatial range. As a result, the network can exploit dense geometric structure for nearby objects and maintain semantic consistency for distant regions where LiDAR evidence is weak. This 2D setting complements rather than replaces 3D perception: it targets stacks that already contain both sensors and require dense camera-aligned labels, rather than advocating that LiDAR be added solely for 2D segmentation. Our main contributions are summarized as follows:
}

\begin{itemize}[nosep]
    \item {We study LiDAR-assisted RGB image-plane semantic segmentation and propose a range-aware heterogeneous fusion framework for this task.}
    \item {We introduce a C2P projection module that transfers pretrained cylindrical 3D semantic features into 2D RGB segmentation, preserving richer geometric structure than conventional depth-map-based fusion.}
    \item {We develop a RAHF module that decomposes near-field and far-field interaction strategies, improving segmentation accuracy and robustness under challenging weather, illumination, and sensor conditions.}
\end{itemize}

\section{Related Work}
\label{sec:related}

{
\paragraph{Semantic segmentation for autonomous driving.}
Semantic segmentation has developed from fully convolutional networks and encoder-decoder architectures~\cite{long2015fully,ronneberger2015u} to transformer-based designs that model longer-range context and stronger mask representations~\cite{cheng2022masked,xie2021segformer,liu2021swin}. These advances have substantially improved dense prediction in natural images and driving scenes. However, autonomous driving presents additional difficulties that are not fully captured by generic segmentation benchmarks. Outdoor scenes contain large depth variation, severe class imbalance, and safety-critical small objects. Their appearance also changes dramatically under night-time illumination, over-exposure, rain, fog, shadows, and motion blur~\cite{9729634,zhou2019automated,zhang2026study}. As a result, a model that performs well with clean RGB appearance may still fail when the visual signal is degraded or ambiguous.

\paragraph{Task scope: LiDAR-assisted RGB segmentation.}
LiDAR-camera fusion has been studied under several different output spaces. A large portion of recent work focuses on 3D perception, such as using images to enrich point clouds for 3D semantic segmentation, 3D scene understanding, or object-level prediction~\cite{vora2020pointpainting,li2023mseg3d,kong2025multi}. These tasks are important, but they differ from the setting of this paper. We study image-plane semantic segmentation: the input includes RGB and LiDAR, but the target is a dense 2D label map in the camera view. This distinction matters because methods designed for 3D point labels or 3D boxes cannot be compared directly under the same annotation protocol, output resolution, or evaluation metric. Therefore, our comparisons emphasize RGB-only 2D segmentation models and task-matched RGB-X/RGB-LiDAR 2D fusion baselines, with CMX serving as a particularly relevant reference because it addresses cross-modal semantic segmentation in the image plane.
Table~\ref{tab:task_scope} summarizes this task distinction and explains why we prioritize image-plane segmentation baselines over multimodal methods whose outputs are defined in 3D space.

\begin{table}[t]
\centering
\footnotesize
\setlength{\tabcolsep}{3pt}
\renewcommand{\arraystretch}{1.05}
\resizebox{\columnwidth}{!}{%
\begin{tabular}{@{}lll@{}}
\toprule
Direction & Output space & Matched \\
\midrule
RGB-only 2D segmentation~\cite{long2015fully,cheng2022masked,xie2021segformer} & Image labels & Yes \\
RGB-X / RGB-LiDAR 2D fusion~\cite{hazirbas2016fusenet,wang2022multimodal,zhang2023cmx} & Image labels & Yes \\
Image-assisted 3D perception~\cite{vora2020pointpainting,li2023mseg3d,kong2025multi} & Points/objects & No \\
LiDAR-only 3D segmentation~\cite{zhu2021cylindrical,ando2023rangevit,kong2023rethinking} & Point labels & No \\
LiDAR-assisted 2D segmentation (\textbf{Ours}) & Image labels & \textbf{Yes} \\
\bottomrule
\end{tabular}%
}
\caption{Task-scope taxonomy for LiDAR-camera perception. The ``Matched'' column indicates whether a method predicts labels in our target image space.}
\label{tab:task_scope}
\end{table}

\paragraph{RGB-LiDAR multimodal fusion.}
RGB-LiDAR fusion addresses this limitation by combining dense image appearance with explicit geometric measurements. Existing methods differ mainly in the stage and manner of cross-modal interaction. Early fusion strategies commonly project depth or point measurements into the image plane and concatenate them with RGB features~\cite{hazirbas2016fusenet,8917447}. Sequential or painting-style methods enrich points or pixels with features from the other modality before downstream prediction~\cite{vora2020pointpainting}. More recent approaches use token exchange, cross-modal attention, or multi-resolution fusion to improve feature interaction~\cite{wang2022multimodal,zhang2023cmx,broedermann2023hrfuser}. These designs demonstrate that LiDAR can improve perception when the two modalities are aligned and complementary. However, most of them still treat the projected LiDAR representation as a spatially homogeneous auxiliary signal. The same convolution, attention, or token-mixing rule is usually applied to nearby and distant regions, even though the LiDAR point density and reliability are very different across range. More generally, representation-level multimodal learning also investigates how to preserve modality-specific diversity while controlling excessive cross-modal disagreement~\cite{xia2026dagr}.

\paragraph{3D LiDAR representation learning.}
In parallel with multimodal fusion, LiDAR representation learning has rapidly advanced~\cite{lin2025hilots}. Point-based methods operate directly on unordered point sets~\cite{qi2017pointnet,guo2020deep}, while pillar-, voxel-, cylindrical-, and range-view-based methods organize point clouds into structured forms for efficient learning~\cite{lang2019pointpillars,zhu2021cylindrical,ando2023rangevit,kong2023rethinking}. Cylindrical voxelization is particularly suitable for driving scenes because it follows the radial distribution of LiDAR scans and better balances near- and far-range spatial resolution~\cite{zhu2021cylindrical,zhou2020cylinder3d}. Compared with raw depth maps, high-level 3D features extracted by such encoders contain richer semantic shape, local geometry, and density information. Yet many 2D RGB-LiDAR segmentation pipelines only use shallow projected depth or range cues, leaving the representation power of pretrained 3D encoders underexploited.

\paragraph{Cross-view projection and feature alignment.}
Bridging 3D LiDAR features and 2D image features requires cross-view alignment. A straightforward strategy is to project raw points or range-view measurements into the image plane, which is computationally simple but discards much of the original 3D feature structure~\cite{8917447,vora2020pointpainting,zhang2023cmx}. Cross-view feature projection provides a stronger alternative: learned 3D features can be associated with their physical coordinates and then injected into corresponding image locations through camera calibration. This process must handle sparsity, occlusion, many-to-one projection conflicts, and resolution mismatch between 3D voxel grids and 2D feature maps. Our C2P module follows this direction by projecting non-empty cylindrical voxel features into the perspective image plane, using pooling to resolve collisions and channel alignment to make the projected representation compatible with the RGB branch. This allows 2D segmentation to benefit from structured 3D priors without replacing the efficient image decoder.

% \paragraph{Range-aware robustness in driving scenes.}
% Robust perception has become increasingly important as autonomous systems are evaluated under corruption, adverse weather, and long-tail scenarios~\cite{kong2023rethinking,zhang2023deliveringing}. For LiDAR-camera fusion, robustness is not only a matter of adding more modalities; the reliability of each modality changes with scene conditions and spatial range. LiDAR can stabilize perception when RGB appearance is degraded, but sparse or misprojected far-field points can also harm semantic consistency. Conversely, RGB context is useful for distant regions, but it may fail around nearby object boundaries under shadows or exposure artifacts. These observations motivate adaptive fusion mechanisms that do not impose a single interaction rule over the entire image. In contrast to uniform fusion, \method\ explicitly models range-dependent modality reliability: it uses LiDAR-guided geometric enhancement in the near field and visual-context enhancement in the far field, with a soft mask providing a smooth transition between the two regimes.
}

\section{Methods}
\label{sec:method}

\begin{figure*}[t]
    \centering
    \includegraphics[width=\linewidth]{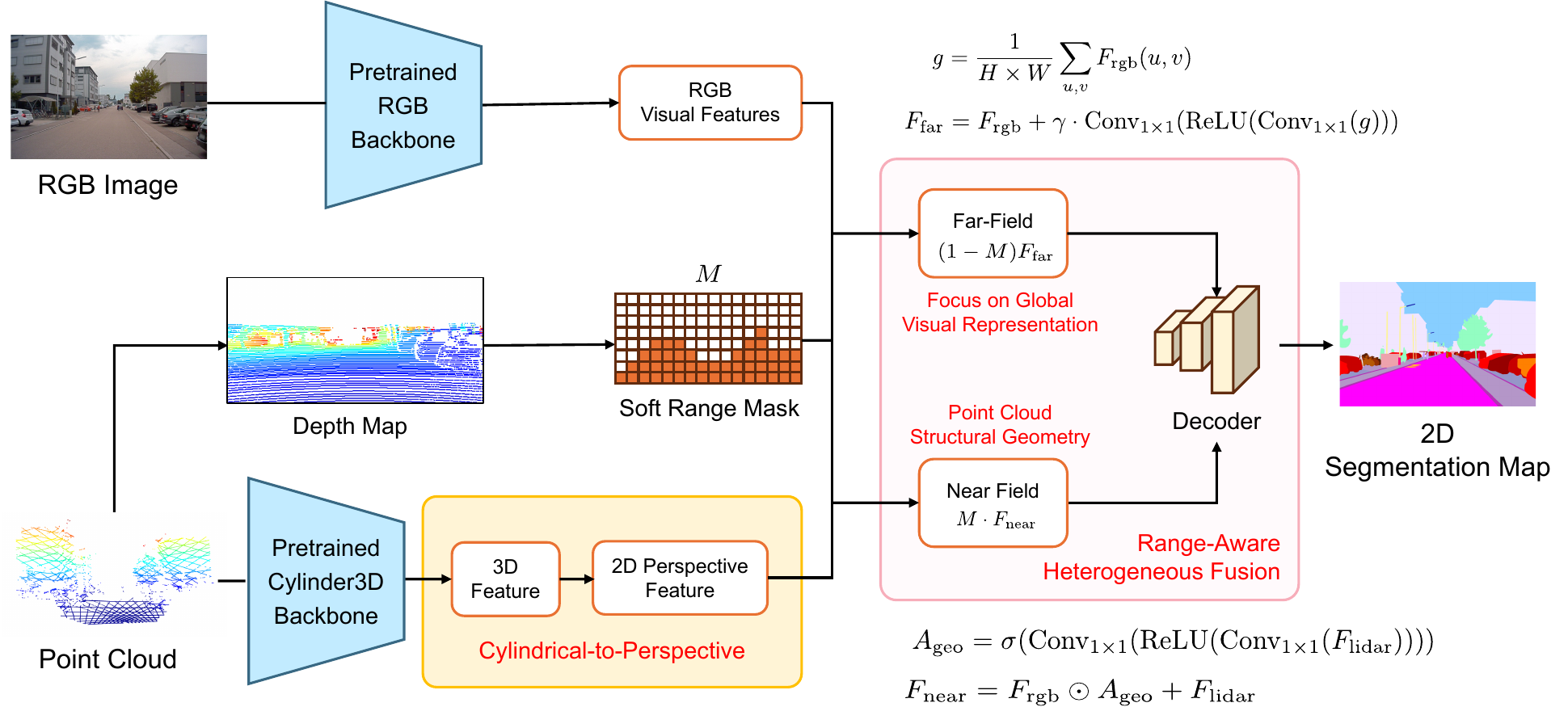}
    \caption{{Overview of \method. The proposed C2P module projects pretrained cylindrical LiDAR features into the image plane, and the RAHF module performs range-aware fusion with near-field geometric refinement and far-field visual context reasoning.}}
    \label{fig:pipeline}
\end{figure*}

{
\subsection{Overview}

Figure~\ref{fig:pipeline} presents the overall pipeline of \method. Given an RGB image and the corresponding LiDAR point cloud, the network predicts a dense semantic map in the RGB image plane. The LiDAR stream is used as an auxiliary source of 3D geometric and semantic priors, not as a separate point-wise prediction target. The network follows four stages. First, an asymmetric dual-stream encoder extracts dense 2D visual features and sparse 3D cylindrical LiDAR features. The RGB branch captures appearance, texture, and scene context, while the LiDAR branch preserves geometric structure in the original 3D space. Second, the proposed Cylindrical-to-Perspective (C2P) module projects non-empty cylindrical voxel features into the perspective image plane and aligns them with the image feature map. Third, a soft range mask is generated from the projected LiDAR depth, providing a continuous estimate of whether each feature location belongs to the near field or far field. Finally, the Range-Aware Heterogeneous Fusion (RAHF) module performs different interactions in different regions: LiDAR-guided geometric enhancement for reliable near-field points and visual-context enhancement for sparse far-field regions. The fused representation is decoded into dense image-plane semantic predictions.

The full inference procedure is summarized in Algorithm~\ref{alg:rahf}. The algorithm highlights that C2P and RAHF are not independent heuristics: C2P supplies structured 3D priors in the 2D plane, while the range mask determines where those priors should dominate and where visual context should take over.
}

\begin{algorithm}[t]
\caption{{Algorithmic workflow of \method.}}
\label{alg:rahf}
\begin{algorithmic}[1]

\Require RGB image $I$, LiDAR point cloud $P$, camera intrinsic matrix $K$, LiDAR-to-camera extrinsic matrix $T$
\Ensure Pixel-wise semantic prediction $\hat{Y}$
\State Extract RGB features $\mrgb \leftarrow E_{\mathrm{img}}(I)$.
\State Voxelize $P$ in cylindrical coordinates and extract 3D features $\vcyl \leftarrow E_{\mathrm{cyl}}(\mathrm{Voxelize}(P))$.
\State Recover non-empty cylindrical voxel centers and form point-feature pairs $\{(P_k,f_k)\}$.
\State Project each $P_k$ to image coordinates $(u_k,v_k)$ using $K$ and $T$.
\State Construct projected LiDAR feature map $\mlidar^{\mathrm{proj}}$ by assigning $f_k$ to $(u_k,v_k)$ with max pooling for collisions.
\State Downsample and channel-align $\mlidar^{\mathrm{proj}}$ to obtain $\mlidar$.
\State Project raw LiDAR points to build sparse depth map $D$ with min pooling for visible-surface depth.
\State Generate soft range mask $\maskm$ from normalized depth.
\State Compute near-field feature $\mnear$ with LiDAR-guided geometric attention.
\State Compute far-field feature $\mfar$ with global visual context enhancement.
\State Fuse features by $\mfused \leftarrow \maskm \odot \mnear + (1-\maskm)\odot \mfar$.
\State Decode $\mfused$ to obtain semantic prediction $\hat{Y}$.
\end{algorithmic}
\end{algorithm}

{
\subsection{Asymmetric Dual-Stream Feature Extraction}

For the RGB branch, given an input image $I \in \mathbb{R}^{H \times W \times 3}$, we use a standard 2D backbone to extract semantic visual features:
\begin{equation}
\mrgb = E_{\mathrm{img}}(I) \in \mathbb{R}^{\frac{H}{s}\times \frac{W}{s}\times C},
\label{eq:rgb}
\end{equation}
where $s$ denotes the spatial downsampling ratio and $C$ is the feature channel dimension. These features retain dense image appearance, local texture, and high-level semantic context, which are essential for classifying distant or texture-dominated regions.

For the LiDAR branch, instead of projecting the point cloud into a shallow depth map before representation learning, we use a pretrained cylindrical voxel encoder~\cite{zhu2021cylindrical}. Given a point cloud
\begin{equation}
P = \{p_i\}_{i=1}^{N}, \quad p_i=(x_i,y_i,z_i,r_i),
\end{equation}
we first transform points into cylindrical coordinates and divide them into a sparse cylindrical grid along radius, azimuth, and height. The 3D encoder then produces high-level cylindrical voxel features:
\begin{equation}
\vcyl = E_{\mathrm{cyl}}(\mathrm{Voxelize}(P)) \in \mathbb{R}^{C' \times R \times \Theta \times H'},
\label{eq:cyl}
\end{equation}
where $C'$ is the 3D feature dimension, and $R$, $\Theta$, and $H'$ denote the radial, angular, and vertical grid sizes. Cylindrical voxelization is well matched to the radial sampling pattern of outdoor LiDAR scans. Therefore, $\vcyl$ preserves not only local geometry but also range-dependent density patterns that would be difficult to recover from a 2D depth projection.

\subsection{Cylindrical-to-Perspective 3D Feature Projection}

The 3D cylindrical features $\vcyl$ cannot be directly fused with image features because the two representations are defined in different coordinate systems. We therefore introduce C2P to convert sparse cylindrical voxel features into a 2D feature map aligned with $\mrgb$.

We first extract all non-empty cylindrical voxels and recover their physical centers in the LiDAR coordinate system. For a voxel indexed by $(i_\rho, i_\theta, i_z)$, its Cartesian coordinates are obtained as
\begin{equation}
x = \rho_i \cos(\theta_i), \qquad
y = \rho_i \sin(\theta_i), \qquad
z = z_i.
\label{eq:cartesian}
\end{equation}
Each valid voxel is then represented as a point-feature pair $(P_k, f_k)$, where $P_k$ denotes the physical 3D location and $f_k \in \mathbb{R}^{C'}$ denotes the corresponding semantic feature vector. Unlike projecting raw points, this operation projects features that have already encoded local 3D context through the pretrained LiDAR encoder.

Using the camera intrinsic matrix $K$ and the LiDAR-to-camera extrinsic matrix $T$, each voxel center $P_k$ is projected into the image plane:
\begin{equation}
z_c
\begin{bmatrix}
u_k\\
v_k\\
1
\end{bmatrix}
=
KT
\begin{bmatrix}
P_k\\
1
\end{bmatrix}.
\label{eq:projection}
\end{equation}
We construct a sparse projected feature map $\mlidar^{\mathrm{proj}}\in\mathbb{R}^{H\times W\times C'}$ by assigning $f_k$ to pixel $(u_k,v_k)$. If multiple 3D voxels are projected to the same pixel, max pooling is used to retain the strongest feature response. Pixels without projected voxels are filled with zeros. This design preserves the sparsity pattern of LiDAR observations while allowing the 2D branch to receive high-level 3D semantic information. Finally, $\mlidar^{\mathrm{proj}}$ is downsampled by the same ratio $s$ as the RGB feature map and passed through a $1\times1$ convolution to align the channel dimension, yielding $\mlidar \in \mathbb{R}^{\frac{H}{s}\times\frac{W}{s}\times C}$.
}

{
\subsection{Range Mask Generation}

After C2P, the network obtains image-aligned 3D semantic features. The remaining question is how much the model should trust these features at each spatial location. A uniform fusion rule is suboptimal because LiDAR density decreases with distance: near-field points are dense and geometrically accurate, while far-field projections are sparse and may act as noise. We explicitly encode this physical prior with a soft range mask.

We project the raw LiDAR points into the image plane and construct a sparse depth map $D$. When multiple points fall into the same pixel, we use min pooling to keep the nearest visible surface, which follows the occlusion principle of perspective imaging and avoids using hidden background points as supervision for the foreground pixel. After downsampling $D$ to the feature resolution, we normalize it by the maximum valid depth $D_{\max}$ and convert it into a continuous mask:
\begin{equation}
\maskm(u,v)=
\sig\left(
\alpha \left(1-\frac{D(u,v)}{D_{\max}}\right)-\beta
\right),
\label{eq:mask}
\end{equation}
where $\sig(\cdot)$ is the sigmoid function. The depth inversion term makes nearer pixels produce larger mask values. The parameter $\alpha$ controls the smoothness of the near-far transition, while $\beta$ determines the physical threshold of the transition. Thus, $\maskm(u,v)\approx 1$ indicates a near-field location where LiDAR is relatively reliable, and $\maskm(u,v)\approx 0$ indicates a far-field location where visual context should play a stronger role.

\subsection{Range-Aware Heterogeneous Fusion}

Given visual features $\mrgb$, aligned LiDAR features $\mlidar$, and the soft range mask $\maskm$, RAHF uses two specialized branches instead of one shared fusion operator.

\paragraph{Near-field geometric enhancement.}
In near-field regions, LiDAR provides accurate geometric boundaries that are useful for correcting RGB ambiguity caused by shadows, reflections, or exposure artifacts. We therefore use $\mlidar$ as a gating signal to recalibrate image features:
\begin{equation}
A_{\mathrm{geo}}=
\sig\!\left(
\mathrm{Conv}_{1\times1}\big(
\mathrm{ReLU}(
\mathrm{Conv}_{1\times1}(\mlidar))
\big)
\right),
\label{eq:ageo}
\end{equation}
\begin{equation}
\mnear = \mrgb \odot A_{\mathrm{geo}} + \mlidar,
\label{eq:fnear}
\end{equation}
where $A_{\mathrm{geo}}$ is a learned geometric attention map and $\odot$ denotes element-wise multiplication. The attention term suppresses visually noisy responses and emphasizes boundary-related locations, while the residual LiDAR term injects explicit 3D structure.

\paragraph{Far-field visual context enhancement.}
In far-field regions, the projected LiDAR features are often sparse or incomplete. Directly forcing geometric fusion can damage the coherence of RGB semantics. We therefore suppress far-field LiDAR dominance and strengthen visual reasoning through global context:
\begin{equation}
g = \frac{1}{HW}\sum_{u,v}\mrgb(u,v),
\label{eq:global}
\end{equation}
\begin{equation}
\mfar = \mrgb + \gamma \cdot
\mathrm{Conv}_{1\times1}\big(
\mathrm{ReLU}(
\mathrm{Conv}_{1\times1}(g))
\big),
\label{eq:ffar}
\end{equation}
where $g$ is a global visual descriptor and $\gamma$ is a learnable scaling factor. This branch allows the model to infer distant semantics from the overall scene layout when local geometry is weak.

\paragraph{Distance-adaptive combination.}
The final fused representation is obtained by softly combining the two branches:
\begin{equation}
\mfused = \maskm \odot \mnear + (1-\maskm)\odot \mfar.
\label{eq:fused}
\end{equation}
This formulation avoids a hard near-far boundary and keeps feature transitions spatially smooth. Near the ego vehicle, $\mfused$ is dominated by LiDAR geometric enhancement; at long range, it gradually shifts toward context-enhanced RGB reasoning. Since RAHF is implemented with $1\times1$ convolutions, global pooling, and element-wise operations, it shows modest overhead while changing the fusion behavior in a physically meaningful way.

\subsection{Decoder and Training Objective}

The fused feature map $\mfused$ is fed into a standard segmentation decoder that upsamples the representation to the original image resolution and predicts a class distribution for each pixel. We denote the prediction by $\hat{Y}\in\mathbb{R}^{H\times W\times N_{\mathrm{class}}}$.

Driving-scene segmentation is highly imbalanced: road and building pixels dominate the image, whereas pedestrians, traffic signs, poles, and other safety-critical objects occupy only a small fraction of pixels. To mitigate this issue, we optimize a joint loss with weighted cross-entropy and Lovasz-Softmax~\cite{berman2018lovasz}. The weighted cross-entropy term is
\begin{equation}
\mathcal{L}_{\mathrm{wce}}
=
-\frac{1}{N_{\mathrm{pix}}}
\sum_i w_{c_i}\log \hat{Y}_{i,c_i},
\label{eq:wce}
\end{equation}
where $c_i$ is the ground-truth class of pixel $i$, and $w_{c_i}$ is the class weight. This term increases the penalty for underrepresented categories. Lovasz-Softmax provides an IoU-oriented surrogate that better matches the final evaluation metric. The total loss is
\begin{equation}
\mathcal{L}_{\mathrm{total}}
=
\mathcal{L}_{\mathrm{wce}}
+
\lambda \mathcal{L}_{\mathrm{lovasz}},
\label{eq:loss}
\end{equation}
where $\lambda$ is set to $1.0$ in all experiments. The two terms are complementary: weighted cross-entropy stabilizes pixel-wise classification under long-tail class distributions, while Lovasz-Softmax directly encourages improvements in mIoU.
}

\paragraph{Interpretation and cost.}
The depth-derived mask is a transparent routing prior, not a learned uncertainty estimator; both branches remain learned end to end. This isolates the range prior without adding a confidence network. Although RAHF itself is lightweight, the full system inherits the LiDAR encoder's cost, so we make no end-to-end real-time claim.

\section{Experiments}
\label{sec:exp}

\subsection{Datasets and Metrics}

{
We use KITTI-360~\cite{liao2022kitti} as the main benchmark for RGB-LiDAR semantic segmentation, and report additional robustness analysis on DELIVER~\cite{zhang2023delivering} in Section~\ref{sec:robustness}. The main task studied in this paper is 2D pixel-wise semantic segmentation with RGB-LiDAR fusion, where LiDAR point-cloud features assist RGB image-plane prediction. Therefore, we choose benchmarks that provide synchronized camera images, LiDAR scans, reliable calibration, and dense 2D semantic annotations. We do not compare against single-modal 3D point-cloud segmentation, 3D detection, or scene-completion methods, because their prediction spaces and annotation protocols are different from the 2D image-plane task considered here.

% KITTI-360 is a large-scale urban driving dataset collected in Karlsruhe. It contains synchronized high-resolution RGB images, Velodyne HDL-64E LiDAR point clouds, camera calibration, and semantic annotations for complex street scenes. It is well suited for evaluating C2P because accurate camera-LiDAR calibration is required to project 3D features into the image plane. It is also suitable for range-aware fusion because the scenes include both near-field objects with dense LiDAR support and far-field regions with sparse points.

Following common practice in semantic segmentation, we report mean Intersection-over-Union (mIoU) as the primary metric. In addition to the overall mIoU, we report representative per-class IoUs for road, building, traffic sign, and pedestrian. Road and building are large background categories that dominate driving scenes, while traffic sign and pedestrian are small foreground categories with strong long-tail characteristics and high safety relevance. Reporting these classes allows us to analyze whether a method improves only dominant scene regions or also benefits small objects that are easily missed in image-plane segmentation. Our baseline set is therefore task-matched rather than simply year-matched: it includes RGB-only 2D segmentation models and representative RGB-X/RGB-LiDAR 2D fusion methods that produce image-plane semantic predictions under the same metric.
}

% We evaluate the proposed method on KITTI-360~\cite{liao2022kitti} as the main benchmark for RGB-LiDAR semantic segmentation. Following standard practice, we report mean Intersection-over-Union (mIoU) as the main metric, together with selected per-class IoUs for representative foreground and background categories. Additional robustness evaluation on DELIVER~\cite{zhang2023delivering} is provided in the appendix.

\subsection{Implementation Details}

{
For the visual branch, we adopt SegFormer-B2~\cite{xie2021segformer} as the 2D backbone because it provides a strong balance between local detail and global context. For the LiDAR branch, we use the pretrained cylindrical encoder proposed in Cylinder3D~\cite{zhu2021cylindrical}. The cylindrical grid configuration follows the pretrained model, with a maximum radial distance of $50$m, full $360^\circ$ azimuth coverage, vertical range of $(-4,2)$, and voxel resolution of $(240,180,20)$. Both projected LiDAR features and image features are aligned to $256$ channels before fusion. The segmentation decoder upsamples the fused feature map to the original image resolution and produces per-pixel class probabilities.

The model is trained with AdamW~\cite{loshchilov2017decoupled}, using an initial learning rate of $6\times10^{-5}$, weight decay of $0.01$, and a polynomial decay schedule with power $0.9$. All experiments are conducted on four NVIDIA RTX 3090 GPUs with a batch size of $4$ per GPU. For the range mask, we set $\alpha=10.0$ and $\beta=5.0$ according to the ablation in Table~\ref{tab:ablation_hyper}. Unless otherwise specified, all reported KITTI-360 results use the same training protocol and backbone configuration, so that the comparison mainly reflects the effect of the fusion design.
}

\subsection{Main Results on KITTI-360}

% Table~\ref{tab:main} compares \method\ against representative RGB-only and RGB-LiDAR baselines on KITTI-360. Among RGB-only models, our method significantly outperforms Mask2Former~\cite{cheng2022masked}, indicating the benefit of complementary LiDAR geometry in challenging outdoor scenes. More importantly, compared with RGB-LiDAR fusion baselines such as HRFuser~\cite{broedermann2023hrfuser}, TokenFusion~\cite{wang2022multimodal}, and CMX~\cite{zhang2023cmx}, \method\ achieves the best overall performance with 65.76 mIoU, surpassing CMX by 1.45 points.

{
Table~\ref{tab:main} compares \method\ with task-matched RGB-only and RGB-LiDAR methods that output 2D semantic maps. \method\ achieves the best result with 65.76 mIoU, exceeding the strongest RGB-only baseline, Mask2Former, by 9.39 points and the strongest fusion baseline, CMX, by 1.45 points. The improvement over CMX suggests that the gain is not explained by generic cross-modal interaction alone. It extends to both dominant background classes and small safety-critical classes, supporting the use of complementary geometry rather than RGB appearance alone.
}

\begin{table}[t]
\centering
\scriptsize
\setlength{\tabcolsep}{2.2pt}
\renewcommand{\arraystretch}{1.05}
\resizebox{\columnwidth}{!}{%
\begin{tabular}{@{}lcccccc@{}}
\toprule
Method & Mod. & mIoU & Road & Building & T. Sign & Ped. \\
\midrule
FCN~\cite{long2015fully} & R & 54.02 & 91.44 & 80.29 & 44.36 & 51.62 \\
U-Net~\cite{ronneberger2015u} & R & 54.83 & 91.73 & 81.36 & 44.79 & 51.48 \\
Mask2Former~\cite{cheng2022masked} & R & 56.37 & 93.52 & 83.69 & 45.13 & 52.93 \\
HRFuser~\cite{broedermann2023hrfuser} & R+L & 48.74 & 87.61 & 78.81 & 37.24 & 48.51 \\
TokenFusion~\cite{wang2022multimodal} & R+L & 54.55 & 92.18 & 82.25 & 43.94 & 52.47 \\
CMX~\cite{zhang2023cmx} & R+L & 64.31 & 94.57 & 87.43 & 52.08 & 58.24 \\
\textbf{\method} & R+L & \textbf{65.76} & \textbf{94.92} & \textbf{88.78} & \textbf{52.22} & \textbf{59.96} \\
\bottomrule
\end{tabular}%
}
\caption{Comparison with task-matched methods on KITTI-360. R and L denote RGB and LiDAR, respectively.}
\label{tab:main}
\end{table}

{
Simply adding LiDAR is not sufficient: HRFuser and TokenFusion remain below the strongest RGB-only baseline in this setting, consistent with the difficulty of handling projected sparsity and feature conflict. By contrast, \method\ reaches 94.92 and 88.78 IoU on road and building, and 52.22 and 59.96 on traffic sign and pedestrian. Thus, the improvement is not confined to easy background pixels; it also covers small foreground regions that often occur at long range or near cluttered boundaries.

% These observations support the central motivation of this paper: RGB-LiDAR segmentation should not be treated as a uniform feature-mixing problem. C2P injects structured 3D cylindrical features into the image plane, while RAHF determines where LiDAR should dominate and where visual context should be trusted more. The combination allows reliable near-field geometry to refine object boundaries without forcing sparse far-field LiDAR cues to dominate distant visual semantics.
}

\begin{figure*}[t]
    \centering
    \includegraphics[width=0.82\textwidth]{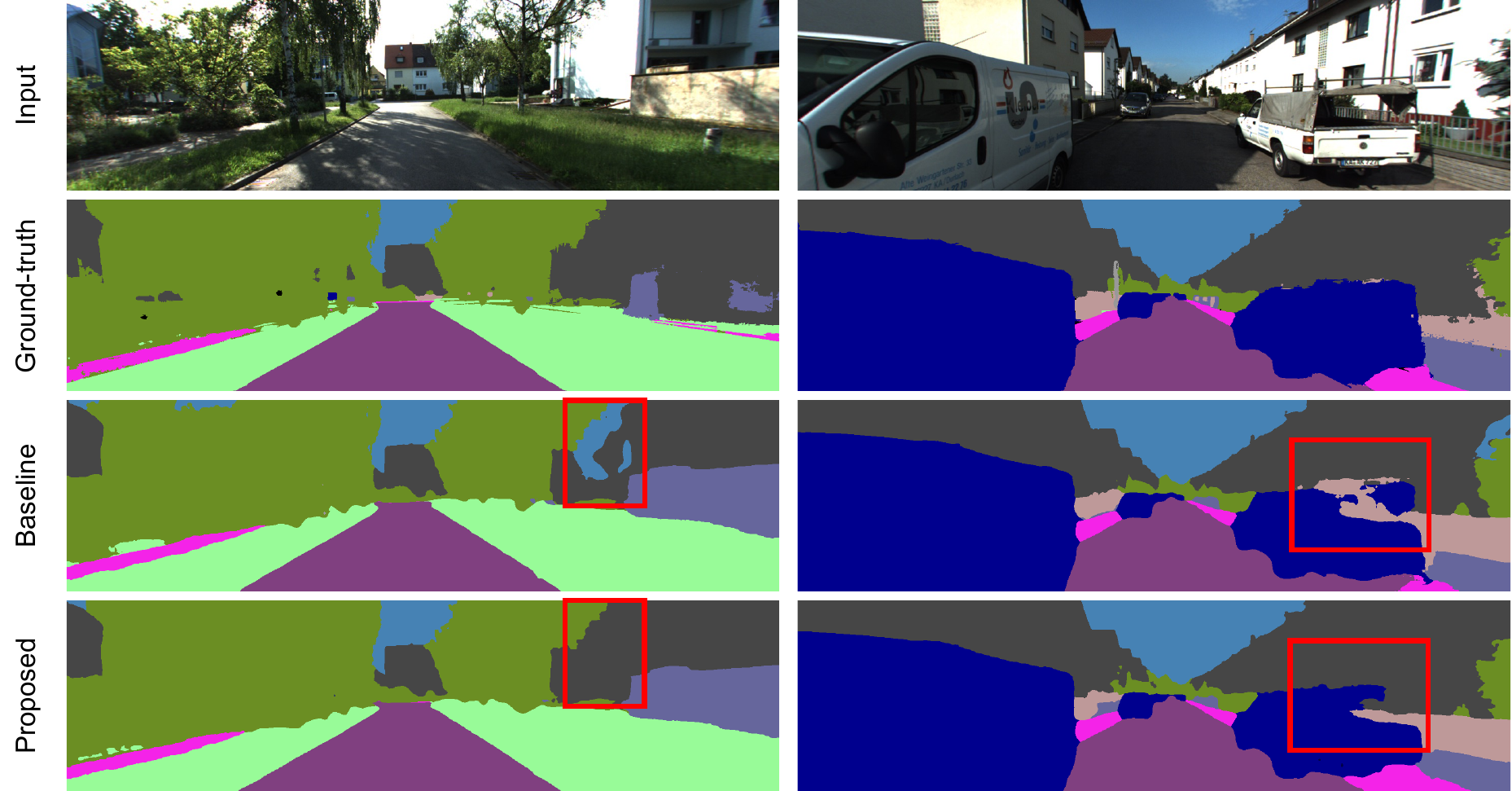}
    \caption{Qualitative comparison on KITTI-360. Red boxes highlight improvements in a distant building and a nearby vehicle.}
    \label{fig:qualitative}
\end{figure*}

Figure~\ref{fig:qualitative} provides complementary qualitative evidence. In the left example, the baseline confuses part of a distant, strongly illuminated building with the sky, whereas \method\ preserves a more complete building region. In the right example, \method\ produces a cleaner boundary for the nearby vehicle. These examples are consistent with the intended behavior of RAHF: visual context stabilizes ambiguous far-field semantics, while denser near-field geometry helps refine object boundaries. We treat them as illustrative rather than as a substitute for the quantitative evaluation.

\subsection{Ablation Study}

% \paragraph{Effect of 3D feature representation.}
% We first investigate the effect of different LiDAR representations. Replacing the pretrained cylindrical feature with a 2D range-view projection yields only a limited improvement, suggesting that simple depth compression loses important structural information. Cartesian voxelization performs slightly better, but still remains inferior to the pretrained cylindrical representation. The best result is achieved by the proposed cylindrical 3D feature injection, verifying that preserving structured 3D priors is beneficial for 2D semantic segmentation.

\paragraph{Effect of 3D feature representation.}
We first study the effect of different LiDAR representations. As shown in Table~\ref{tab:ablation_feature}, adding a 2D range projection improves the 2D-only baseline from 62.18 to 63.12 mIoU, showing that shallow geometric cues are helpful but limited. Cartesian voxelization further improves the result to 63.42 mIoU by retaining more 3D structure. The pretrained cylindrical encoder achieves the best performance with 65.76 mIoU, confirming that cylindrical voxel features better match the radial distribution and range-dependent density of driving-scene LiDAR. These results validate the motivation of C2P: transferring structured 3D features to the image plane is more effective than using shallow projected geometry.

\begin{table}[t]
\centering
\scriptsize
\setlength{\tabcolsep}{2.2pt}
\renewcommand{\arraystretch}{1.05}
\resizebox{\columnwidth}{!}{%
\begin{tabular}{@{}lccccc@{}}
\toprule
3D feature & mIoU & Road & Building & T. Sign & Ped. \\
\midrule
Without 3D feature & 62.18 & 92.72 & 87.05 & 47.51 & 55.74 \\
2D range projection & 63.12 & 94.04 & 86.78 & 48.12 & 56.53 \\
Cartesian voxelization & 63.42 & 93.15 & 87.91 & 48.40 & 56.85 \\
Pretrained cylindrical encoder & \textbf{65.76} & \textbf{94.92} & \textbf{88.78} & \textbf{52.22} & \textbf{59.96} \\
\bottomrule
\end{tabular}%
}
\caption{Ablation on LiDAR feature representations.}
\label{tab:ablation_feature}
\end{table}
% \vspace{-8mm}

\paragraph{Effect of soft mask hyperparameters.}
% We first study the effect of different LiDAR representations. As shown in Table~\ref{tab:ablation_feature}, adding a 2D range projection improves the 2D-only baseline from 62.18 to 63.12 mIoU, showing that shallow geometric cues are helpful but limited. Cartesian voxelization further improves the result to 63.42 mIoU by retaining more 3D structure. The pretrained cylindrical encoder achieves the best performance with 65.76 mIoU, confirming that cylindrical voxel features better match the radial distribution and range-dependent density of driving-scene LiDAR. These results validate the motivation of C2P: transferring structured 3D features to the image plane is more effective than using shallow projected geometry.

To further analyze the impact of hyperparameters $\alpha$ and $\beta$ on segmentation performance during RAHF, we conduct a grid search ablation study with $\alpha\in \{5, 10, 20\}$ and $\beta \in \{2, 5, 8\}$. The overall mIoU results on the KITTI-360 validation set are presented in Table \ref{tab:ablation_hyper}. The results indicate that the model achieves an optimal mIoU of 65.76\% when $\alpha=10$ and $\beta=5$. This implies that the best performance is obtained when the near-far transition threshold is set at approximately 50\% of the maximum detection range with moderate smoothness. 
As shown in Equation \ref{eq:mask}, $\alpha$ controls the smoothness of the mask. 
% If $\alpha$ is too large, the mask degenerates into a hard cut-off operation, disrupting the feature coherence of individual objects in the transition region. Conversely, if $\alpha$ is too small, the transition area loses its distinctiveness, making it difficult to differentiate between near and far features. 
% Both extremes lead to a degradation in segmentation accuracy.
Meanwhile, $\beta$ controls the physical critical threshold separating the near and far regions. An excessively large or small $\beta$ results in either the introduction of far-field noise or the under-utilization of near-field geometric features, which inevitably limits the performance.

\begin{table}[t]
\centering
\footnotesize
\setlength{\tabcolsep}{5pt}
\renewcommand{\arraystretch}{1.05}
\begin{tabular}{@{}lccc@{}}
\toprule
 & $\beta=2$ & $\beta=5$ & $\beta=8$ \\
\midrule
$\alpha=5$ & 64.74 & 62.81 & 62.43 \\
$\alpha=10$ & 64.29 & \textbf{65.76} & 64.71 \\
$\alpha=20$ & 63.87 & 64.82 & 65.37 \\
\bottomrule
\end{tabular}
\caption{Ablation on the soft-mask hyperparameters.}
\label{tab:ablation_hyper}
\end{table}
% \vspace{-8mm}

\paragraph{Effect of fusion strategy.}
We then compare different fusion strategies. Using only near-field 3D enhancement improves segmentation over the 2D-only baseline, confirming the value of LiDAR geometry for nearby objects. Using only far-field context yields a smaller improvement, while applying 3D fusion uniformly across near and far regions does not perform as well. The best result is obtained by the proposed gated heterogeneous fusion, showing that explicitly separating near-field and far-field interaction is more effective than applying a shared fusion rule across the whole scene.

\begin{table}[t]
\centering
\scriptsize
\setlength{\tabcolsep}{2.2pt}
\renewcommand{\arraystretch}{1.05}
\resizebox{\columnwidth}{!}{%
\begin{tabular}{@{}lccccc@{}}
\toprule
Fusion strategy & mIoU & Road & Building & T. Sign & Ped. \\
\midrule
2D only & 62.18 & 92.72 & 87.05 & 47.51 & 55.74 \\
Near-field 3D only & 63.61 & 93.56 & 87.47 & 48.92 & 53.89 \\
Far-field context only & 62.39 & 93.18 & 87.19 & 48.24 & 55.20 \\
3D in both near/far fields & 63.15 & 92.90 & 87.30 & 48.66 & 56.77 \\
Gated heterogeneous fusion & \textbf{65.76} & \textbf{94.92} & \textbf{88.78} & \textbf{52.22} & \textbf{59.96} \\
\bottomrule
\end{tabular}%
}
\caption{Ablation on the fusion strategy.}
\label{tab:ablation_fusion}
\end{table}

% \paragraph{Effect of loss function.}
% Finally, we study the effect of different loss configurations. Weighted cross-entropy improves performance over plain cross-entropy by alleviating class imbalance, while Lovasz loss further benefits IoU-oriented optimization. The best result is obtained by combining weighted cross-entropy with Lovasz-Softmax, indicating that the two objectives are complementary in dense multimodal segmentation.

% \begin{table}[h]
% \centering
% \vspace{-5pt}
% \begin{tabular}{lccccc}
% \toprule
% Loss Function & mIoU & Road & Building & T. Sign & Ped. \\
% \midrule
% Cross-entropy & 63.34 & 93.17 & 86.93 & 48.68 & 56.12 \\
% Weighted cross-entropy & 63.95 & 93.35 & 87.04 & 49.32 & 56.31 \\
% Lovasz loss & 64.02 & 94.06 & 87.23 & 51.21 & 57.84 \\
% Cross-entropy + Lovasz & 65.37 & 94.78 & 88.91 & 51.36 & 59.23 \\
% Weighted cross-entropy + Lovasz & \textbf{65.76} & \textbf{94.92} & \textbf{88.78} & \textbf{52.22} & \textbf{59.96} \\
% \bottomrule
% \end{tabular}
% \vspace{2mm}
% \caption{Ablation on the loss function.}
% \label{tab:ablation_loss}
% \end{table}
% \vspace{-5mm}

\paragraph{Robustness under challenging conditions.}
\label{sec:robustness}
{
We further evaluate \method\ on DELIVER~\cite{zhang2023delivering} under adverse weather, illumination variation, and sensor corruption. As shown in Table~\ref{tab:deliver}, \method\ reaches 58.05 mIoU overall, 5.08 points above the available task-matched baseline. The condition-wise results are consistent with the design motivation: projected 3D structure complements degraded RGB appearance, while the range-aware route avoids relying uniformly on geometry under LiDAR jitter. Because only one external baseline is available here, we use DELIVER as a complementary stress test rather than a comprehensive state-of-the-art comparison. Under-exposure remains the hardest condition at 37.22 mIoU, showing that geometry cannot fully compensate for severely degraded visual evidence.
}

\begin{table}[t]
\centering
\scriptsize
\setlength{\tabcolsep}{2.2pt}
\renewcommand{\arraystretch}{1.02}
\resizebox{\columnwidth}{!}{%
\begin{tabular}{@{}lccccc@{}}
\toprule
Condition & mIoU & Road & Building & T. Sign & Ped. \\
\midrule
Cloudy & 61.37 & 98.17 & 89.75 & 59.39 & 62.54 \\
Foggy & 58.78 & 98.39 & 80.36 & 65.85 & 68.17 \\
Night & 51.08 & 97.43 & 71.25 & 59.78 & 63.38 \\
Rainy & 60.11 & 97.86 & 88.87 & 37.64 & 71.98 \\
Sunny & 60.25 & 98.06 & 88.09 & 49.35 & 72.71 \\
Motion blur & 57.91 & 98.04 & 88.19 & 61.83 & 69.87 \\
Over-exposure & 56.23 & 97.49 & 89.19 & 57.97 & 57.54 \\
Under-exposure & 37.22 & 95.11 & 54.69 & 24.68 & 45.38 \\
LiDAR jitter & 58.49 & 98.30 & 84.58 & 43.42 & 71.08 \\
\midrule
Baseline~\cite{broedermann2023hrfuser} & 52.97 & 96.24 & 80.72 & 45.83 & 61.67 \\
Overall & \textbf{58.05} & \textbf{97.98} & \textbf{84.68} & \textbf{51.18} & \textbf{67.68} \\
Gain & +5.08 & +1.74 & +3.96 & +5.35 & +6.01 \\
\bottomrule
\end{tabular}%
}
\caption{Robustness on DELIVER under challenging conditions.}
\label{tab:deliver}
\end{table}

\paragraph{Evaluation scope.}
Aggregate mIoU uses the full label set; the four displayed classes are diagnostic examples. FCN and U-Net are historical controls, while the fusion claim rests on task-matched methods, with CMX the strongest reference in Table~\ref{tab:main}. Since Table~\ref{tab:deliver} has one external baseline, DELIVER is a complementary stress test rather than a comprehensive state-of-the-art ranking.

\section{Conclusion}
\label{sec:conclusion}
We presented \method, a range-aware framework for LiDAR-assisted RGB image-plane segmentation. C2P transfers structured cylindrical features into the camera plane, while RAHF balances near-field geometric refinement and far-field visual context through a soft range prior. KITTI-360 results show gains over task-matched RGB-only and RGB-LiDAR baselines, and DELIVER provides complementary evidence under adverse conditions. These results support range-dependent fusion when LiDAR is already available to assist dense camera-view prediction, while broader datasets and explicit system-level efficiency measurements remain important future work.

{\small
  \bibliographystyle{ieeenat_fullname}
  \bibliography{main}
}
\end{document}